\documentclass[11pt]{article}
\pdfoutput=1

\usepackage{arXiv}

\usepackage[utf8]{inputenc} 
\usepackage[T1]{fontenc}    
\usepackage{hyperref}       
\usepackage{url}            
\usepackage{booktabs}       
\usepackage{amsfonts}       
\usepackage{nicefrac}       
\usepackage{microtype}      
\usepackage{lipsum}
\usepackage{graphicx}
\usepackage{cite}
\usepackage{color}
\usepackage{enumitem}
\usepackage{amsmath}
\usepackage{amssymb}
\usepackage{multicol}
\usepackage{multirow}
\usepackage{algorithm}
\usepackage{algorithmic}
\usepackage{bm}
\usepackage{array}
\usepackage{diagbox}
\usepackage{subfigure}
\usepackage{subfiles}
\usepackage{epstopdf}

\graphicspath{{images/}}

\title{Field Evaluations of A Deep Learning-based Intelligent Spraying Robot with Flow Control for Pear Orchards}

\author{
  Jaehwi Seol $^{1,2}$, Jeongeun Kim$^{1,2}$, and Hyoung Il Son$^{1,2,}$\thanks{Corresponding author.} \\
  1 Department of Rural and Biosystems Engineering, Chonnam National University, \\ Yongbong-ro 77, Gwangju 61186, Republic of Korea\\
  2 Interdisciplinary Program in IT-Bio Convergence System, Chonnam National University,\\ Yongbong-ro 77, Gwangju 61186, Republic of Korea \\
  \texttt{tjfwognl1@gmail.com, vewry12@gmail.com, hison@jnu.ac.kr} \\
}

\begin{document}
\maketitle

\begin{abstract}
This paper proposes a variable flow control system in real time with deep learning using the segmentation of fruit trees in a pear orchard. The flow rate control in real time, undesired pressure fluctuation and theoretical modeling may differ from those in the real world. Therefore, two types of preliminary experiments were designed to examine the linear relationship of the flow rate modeling. Through a preliminary experiment, the parameters of the pulse width modulation (PWM) controller were optimized, and an actual field experiment was conducted to confirm the performance of the variable flow rate control system. As a result of the field experiment, the performance of the proposed system was satisfactory, as it showed that it could reduce pesticide use and the risk of pesticide exposure. Especially, since the field experiment was conducted in an unstructured environment, the proposed variable flow control system is expected to be sufficiently applicable to other orchards.
\end{abstract}


\section{Introduction}
\label{sec:1}

Currently, as research on precision agriculture continues, studies on various agricultural robots to perform precision agriculture are being conducted~\cite{kim2019unmanned}. In agriculture, robots can be used in various applications, and research using unmanned aerial vehicle and unmanned ground vehicles research to increase the efficiency of agricultural work by using swarm robots are being conducted~\cite{kim2020voronoi,ju2019modeling}. In smart farms, agricultural robots are increasingly being used in horticulture. In addition, a variety of harvesting robots ~\cite{bac2014harvesting} and spraying~\cite{kim2020intelligent} are being carried out in the field or orchard environments.

Pesticide spraying is a task that significantly improves the productivity of crops, especially in orchards, where pest control is essential for increasing yield~\cite{gao2018leaf}. However, spraying control requires a large amount of spraying because of the probability of occurrence of pests and irregularities of occurrence. Speed sprayers (SS), which are mainly used in orchards, spray pesticides in all directions without considering the surrounding environment. This type of spraying may cause changes in the soil owing to the abuse of pesticides because many pesticides are indiscriminately controlled, and workers performing the control may be exposed to pesticides and become poisoned~\cite{berenstein2017automatic,guan2015review}. Therefore, there is a need for an intelligent spraying control system that reduces the use of pesticides by spraying an appropriate amount based on the awareness of the environment to obtain stability and economic benefits for workers~\cite{salcedo2020foliar,manandhar2020techno}.

However, during pesticide spray applications using air-assisted sprayers on fruit crops, only a small portion of the total spray volume reaches the target canopy~\cite{escola2013variable}. This type of spray requires a large amount of spray to improve the control performance, which leads to the unnecessary use of pesticides~\cite{chen2012development}. An intelligent spraying system that recognizes the surrounding environment and individual nozzle control according to each environment is required. However, actual orchard environment is not well-organized environment, the surrounding environment must be recognized more robustly. In addition, because a vision is used to distinguish control targets, a system that strongly recognizes light is required, Therefore, this paper proposes a deep learning-based fruit tree recognition system.

For control, the environment needs to be recognized in real time, and a system that calculates and sprays an appropriate control amount based on the acquired image data is required. In addition, while spraying pesticides, only a small part of the total application amount can reach the target point owing to drift. This type of control requires a large amount of spray, resulting in unnecessary pesticide use. Because the on/off method, which recognizes the fruit tree area, maintains a constant flow rate during spraying, selective on/off control cannot be seen as representative of an intelligent control system. Therefore, in a recent research project, a study was conducted to control the amount of spray considering the variability of the canopy~\cite{liu2014development,giles2011smart,berk2016development}.

\subsection{Related Works}

In~\cite{xiao2017intelligent}, an intelligent pesticide spray technique based on a depth-of-field extraction algorithm was developed. The researchers used Microsoft's Kinect camera to calculate the average distance between the camera and a fruit tree, as well as the leaf wall area density. In~\cite{wei2016effects}, a targeting air-assisted sprayer was introduced based on an infrared detection system. Their spraying system could reduce spray volumes and ground deposition, but the spray deposition and coverage of the canopy were also reduced. Furthermore, the target measurement method simply used the depth data and echo of the infrared sensor. This is not suitable for environments with obstacles and other environmental factors.

In~\cite{kim2020intelligent}, the author developed a deep learning-based intelligent control system to control the spray output of individual nozzles in real time. The output of the individual nozzles was sprayed by determining the on/off state using the set threshold value. In~\cite{shen2017development,chen2019control,cai2019design,chen2012development,osterman2013real}, a variable-rate precision sprayer was developed with a high-speed laser scanning sensor to control the spray output of individual nozzles in real time. The spray amount for each nozzle was determined according to the grid volume, and spraying was performed. However, the target tree used in the experiment was artificial and, experiment was not conducted in an actual orchard environment. In~\cite{chen2011development} a Lidar-guided air-assisted variable-rate spraying system was developed for tree and fruit crop applications. The method suggested was not clear as to the manner in which trees are distinguished from other obstacles in an orchard environment. Moreover, the spray on-target experiment was conducted only under laboratory conditions. Additionally, laser and Lidar sensors have the problem of excessive initial costs. 

In~\cite{zhou2018design}, a target spray platform-based deep learning method was introduced. The researchers developed a vision system, spray system, and moving platform for the target sprayer. They used object detection techniques for deep learning-based crop detection. According to the test results, the use of pesticides can be reduced by approximately 46\% by introducing the prevention system developed in this study, and the correct spray action was 99.1\%. Field tests were conducted in a well-organized environment. Practical field experiments should be conducted in an actual environment. Furthermore, they did not elaborate on the deep-learning training in this study.

In~\cite{liu2014development}, a flow control system was developed to achieve various spray outputs based on the structure and movement speed of trees in order to precisely control the flow rate in orchards and nurseries. However, because this was not tested in the field, it is difficult to prove the performance according to the structure of the tree and the moving speed of the platform. In~\cite{butts2019droplet}, the droplet size was controlled to increase the efficiency of the control device and reduce the risk of scattering. The droplet size was determined by pulse width modulation (PWM) control, and optimal PWM was determined by identifying the droplet size distribution and nozzle tip pressure according to duty cycle, nozzle type, and gauge pressure.
 
\subsection{Contributions}
The contributions and novelty of our study are summarized below:

\begin{enumerate}
    \item We proposed a variable flow control in real-time to enhance spraying performance. A variable flow rate control was designed though a preliminary experiment. In the preliminary experiment, the flow rate was modeled and control parameters were determined, and the proposed variable flow rate control system was optimized according to the orchard environment.
    \item We performed a field experiment in a pear orchard to evaluate the variable flow rate control system. The pear orchard was not a well-organized environment for field experiments; therefore, our intelligent spraying system can be applied to other orchards.
    \item We use fruit tree perception, which we utilized in a previous study~\cite{kim2020intelligent}. Various datasets were constructed by acquiring datasets over two years.
\end{enumerate}

\subsection{Structure of Paper}
The remainder of this paper is organized as follows: Section II introduces the intelligent spraying system and fruit tree perception based on deep learning, a method of fusion depth data. In Section III, we present a real-time variable flow rate nozzle control method through a preliminary experiment. In Section IV, we present our experimental setup and results for the verification and evaluation of the proposed real-time variable flow rate nozzle control system in a pear orchard. In Section V, we discuss the experimental results and the challenges for future work. Finally, in Section VI, we summarize the conclusions and provide directions for future research.

\section{Intelligent Spraying system description}

\subsection{Intelligent spraying system}
\begin{figure}[!t]
\begin{center}
\includegraphics[width=10cm]{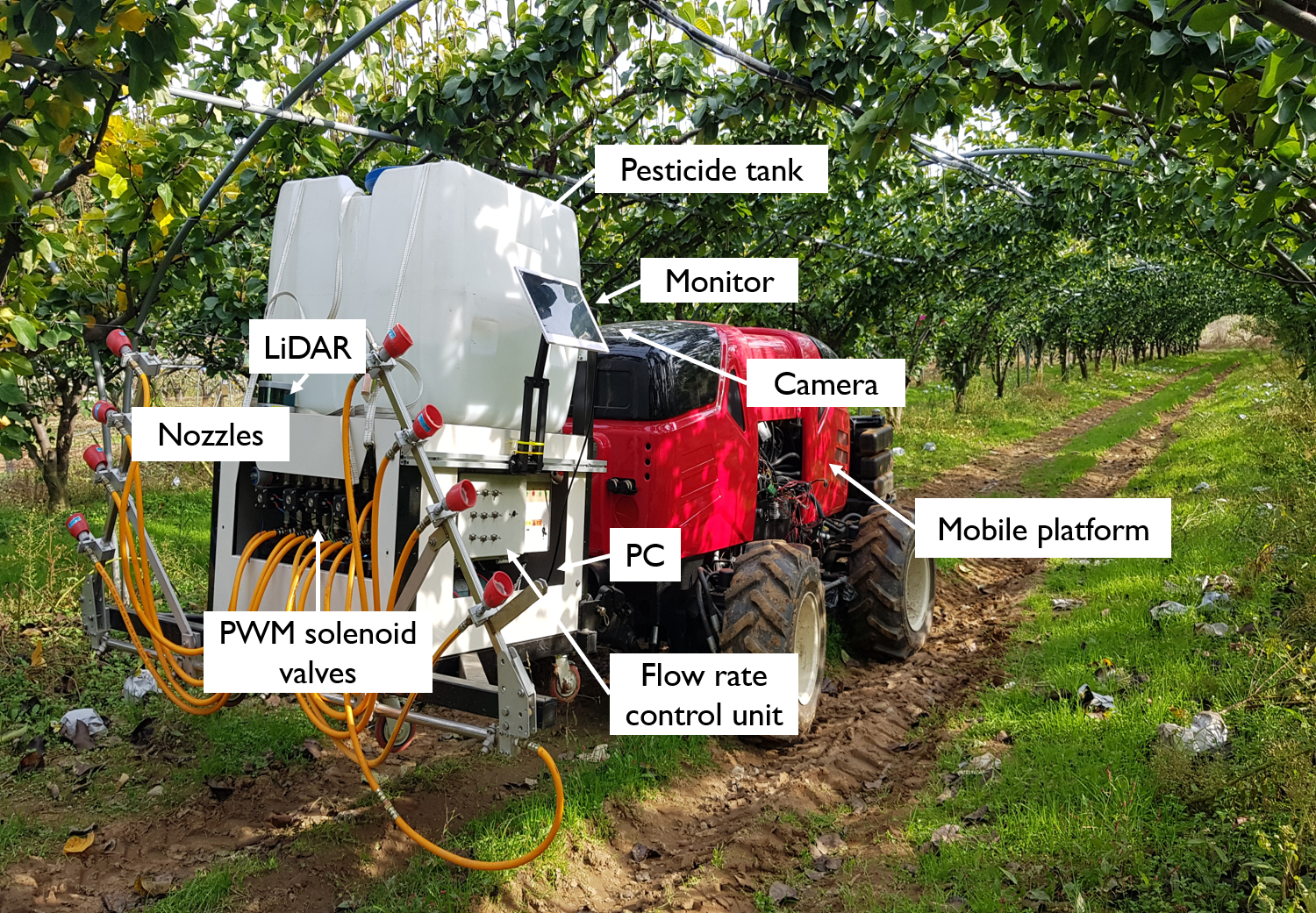}
\caption{Intelligent spraying system. }\label{intelligent spraying}
\end{center}
\end{figure}

The intelligent spraying system is shown in Fig.~\ref{intelligent spraying}. Our spraying system was tightened and lifted through a three-point hitch link on the a mobile platform. The mobile platform can freely drive unstructured roads, such as agricultural fields. Spraying systems and computing platforms receive power from 24 V batteries. The spraying systems were equipped with a 300 L capacity pesticide tank, computing platform, and spray boom with a total of eight nozzles. Two RGB-D cameras were attached to the frame, with one on each side of the platform, and the data were transmitted between the computing platform and the camera.

In order to control each nozzle individually in the orchard, the image obtained from the camera (1280 $\times$ 256) is divided and mapped by considering the location of the nozzle. Because there are four nozzles on one side, the image is divided into four areas and the area is mapped to one nozzle. At this time, the flow rate of each nozzle is determined using deep learning-based fruit tree recognition and depth data with information on the region obtained from the camera.

\subsection{Fruit Tree Perception}

In this study, we recognized fruit trees in orchard environments based on~\cite{kim2020intelligent}. We briefly recapitulate the semantic segmentation for fruit tree perception with depth data fusion to prevent undesired area, and refer to further details in~\cite{kim2020intelligent}.

\subsubsection{Dataset}
The dataset was acquired over a total of two years (2019-2020) from pear orchards using Intel RealSense D435 cameras. Generality was maintained by providing sufficient variation in tree topology, leaf density, and leaf characteristics. A total of 2000 images with a size of 1280$\times$256 were obtained. The fruit of the Korean pear tree is grown in wrapping paper bags to prevent worm and germ infestation and improve quality. The learning process assigns a fruits class to the wrapping paper bags. 

Considering that it is an actual outdoor environment, it is essential to subdivide the classes for deep learning training. There are five classes in the dataset, as acquired from the training: tree (leaf + branch + trunk), fruit, ground, sky, and pipe. The spraying system should only spray areas with fruit and trees, and not in other areas. The components of trees (leaf, branch, and trunk) were grouped into a single class without further division based on details. However, the fruit was grouped into a separate class from the tree because, depending on the season, it may not always be present on the trees. 

\subsubsection{Semantic Segmentation Results}
We previously performed semantic segmentation to perform fruit tree perception. Among the deep learning models commonly used in semantic segmentation, we trained and tested using datasets acquired using U-Net, SegNet, ICNet, and DeepLab v3. The model-specific accuracy is listed in Table~\ref{tab_0}. Perception was performed using the SegNet model with the highest accuracy among the trained models.

\begin{table}[!t]

\centering

\caption{Accuracy performance of segmentation models}
\begin{tabular}{c|cccc} 
\hline
Model & SegNet & U-Net & ICNet & DeepLab v3 \\ 
\hline
Accuracy (\%) & 83.79 & 66.47 & 72.54 & 75.31 \\ 
\hline
\end{tabular}
\label{tab_0}
\end{table}
    
\subsubsection{Fusion Depth Data with Fruit Tree Segmentation}
A pear orchard has trees planted in rows such that the background of one tree can have other trees. Thus, we performed post-processing using the depth data from the RGB-D camera to prevent the trees from being detected in the background. The distance between the camera (which is on the platform) and the tree is approximately 1.2 m, and the total depth of the canopy of the tree, including the branches and leaves, is 0.8 m. The output data from deep learning with more than 2 m of depth data based on the camera are voided. This prevents the trees from being segmented and ensures that only the trees that should be sprayed are segmented.

\section{REAL-TIME VARIABLE FLOW RATE NOZZLE CONTROL}

The flow rate of each nozzle was designed by controlling the installed solenoid valve in real time from the recognized camera image. When controlling each nozzle with this real-time control system, a fluctuation in the spray pressure of the nozzle tip near the solenoid valve mounted near each nozzle tip occurs~\cite{chen2012development}. These pressure fluctuations can cause unintended changes in the flow output and droplet size. These problems can differ greatly from the theoretical modeling of the flow rate and the modeling of the actual flow rate. Therefore, to properly control the solenoid valve, a PWM-based controller is designed, and parameter values for utilizing image information and depth data, which are input data, must be designed and optimized. Therefore, in this study, the variable flow rate control system was experimentally modeled and optimized as follows:

\begin{table}
\caption{Preliminary experiment design}

\begin{tabular}{c|c|cc}
\hline
\multicolumn{2}{c|}{\multirow{2}{*}{\diagbox{Category}{Case}}} & \multirow{2}{*}{Preliminary experiment 1} & \multirow{2}{*}{Preliminary experiment 2} \\
\multicolumn{2}{c|}{} &  &   \\ 
\hline
\multicolumn{2}{c|}{Object} & \begin{tabular}[c]{@{}c@{}}Evaluation of spraying coverage area \\ according to duty cycle \\
to optimize spraying control\end{tabular} & \begin{tabular}[c]{@{}c@{}}Evaluation of spraying distance \\ according to duty cycle \\ to optimize spraying control \end{tabular}  \\ \hline
\multicolumn{2}{c|}{Process} & \begin{tabular}[c]{@{}c@{}} \textit{The fruit area} is changed from 30 to 100\%,\\ \textit{the duty cycle} is changed from 75 to 100\%, \\ and the sprayer is move to forward \end{tabular} & \begin{tabular}[c]{@{}c@{}}\textit{The distance} is changed from 70 to 160cm, \\ \textit{the duty cycle} is changed from 75 to 100\%, \\ and moving the mobile platform with \\artificial tree\end{tabular}  \\ \hline
\multicolumn{2}{c|}{Result} & \begin{tabular}[c]{@{}c@{}}Analysis of spraying coverage area \\ according to duty cycle\\ by pesticide adhesion rate \end{tabular} & \begin{tabular}[c]{@{}c@{}}Analysis of spraying distance \\according to duty cycle \\by pesticide adhesion rate\end{tabular}  \\ \hline
\end{tabular}

\label{tab_pe}
\end{table}

\subsection{Preliminary Experiment}

Although the flow rate changes depending on the PWM duty cycle, the lower the value, the lower the effective spraying coverage area, which is expected to reduce the performance of the spraying. Prior to applying the system to an actual orchard, the design of an appropriate intelligent spraying system is required through preliminary experiments. 

The preliminary experimental results are expressed by the pesticide adhesion rate {$R_p$}, where the pesticide attaches to the water-sensitive paper, and is defined as follows~\cite{kim2020intelligent}:

\begin{eqnarray}
{R}_{p} &=& \frac{A_s}{A_{w}} \times 100 \nonumber\\ &=&\frac{\sum_{i=0}^{r-1}\sum_{j=0}^{c-1}p(i,j)}{r \times c} \times 100
\end{eqnarray}

where {$A_s$} is the sprayed area of the water-sensitive paper,  {$A_w$} is the area of the water-sensitive paper, and r and c are row and column pixels with RGB values corresponding to the sprayed area, respectively.

The preliminary experiments were designed as follows and are shown in Table~\ref{tab_pe}.

\subsubsection{Preliminary Experiment 1: Evaluation of Spraying Coverage Area}
\begin{figure}[!t]
\begin{center}
\includegraphics[width=14cm]{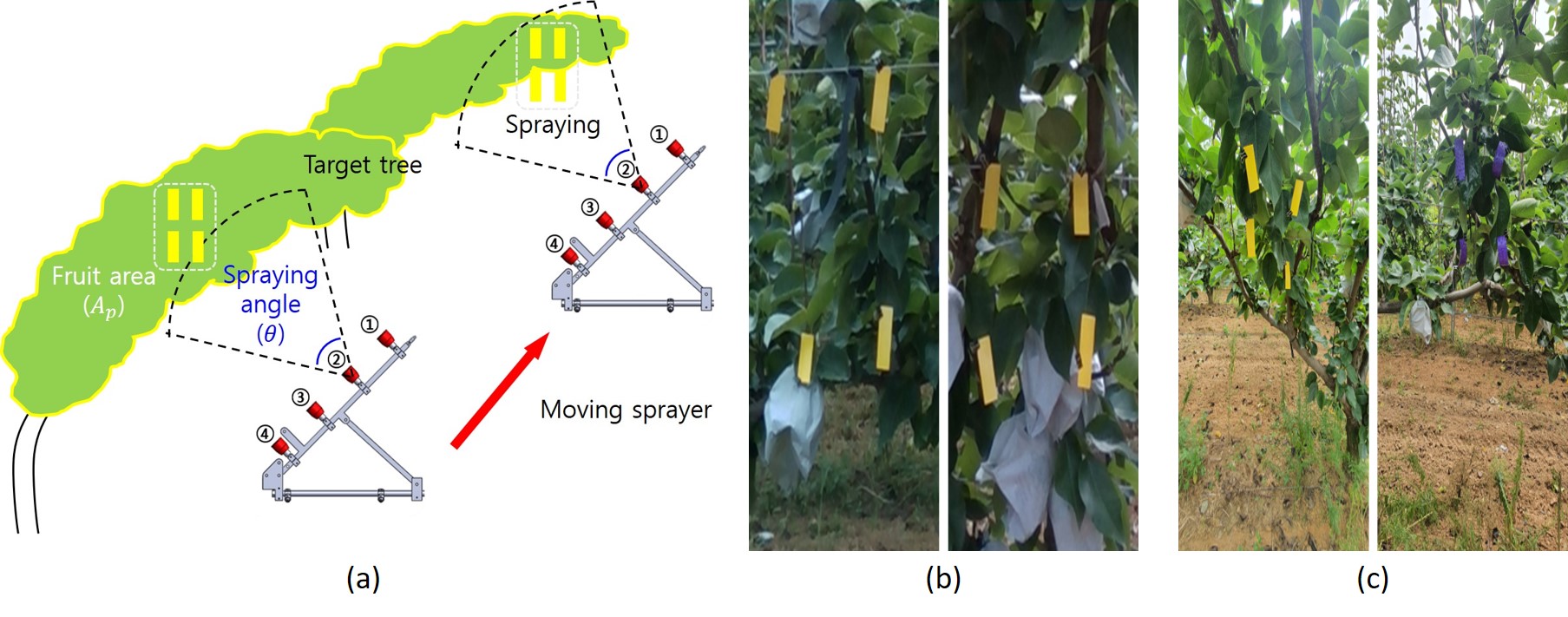}
\end{center}

\caption{Preliminary experiment 1 setup. (a) Configuration of preliminary experiment 1 for spraying performance evaluation. (b) Attached water-sensitive paper according to fruit tree area from 30 to 100\%, 50\% (left) and 90\% (right) (c) Results of spraying: before spraying (left), after spraying (right).}
\label{Preliminary experiment 1}
\end{figure}

As the duty cycle was adjusted, the spray angle decreased. Low duty cycles may not fully cover the desired coverage area. To check whether the spraying was evenly performed in all duty cycles, the duty cycle was set as a variable, and an appropriate duty cycle was optimized by comparing $R_p$. An experiment was designed to find an appropriate duty cycle based on the fruit tree distribution area recognized by the camera. The design of the experiment is shown in Fig.~\ref{Preliminary experiment 1}, and preliminary experiment 1 was conducted on the second nozzle, which showed the most diverse area distribution among the four nozzles. The duty cycle increased from 75\% to 100\%, each fruit tree area was divided into from 30 to 100 in 10\% increments, and four water-sensitive papers were attached to the corresponding area to compare the $R_p$ according to the duty cycle.

\begin{figure}[!t]
\begin{center}
\includegraphics[width=8cm]{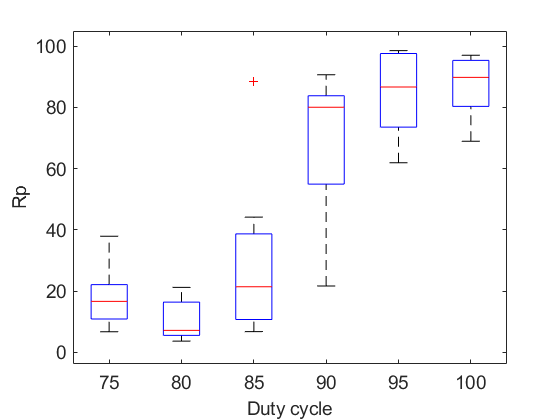}
\caption{Preliminary experiment 1 results. }\label{Pe_re1}
\end{center}
\end{figure}

The results of preliminary experiment 1 are shown in Fig~\ref{Pe_re1}. The experimental results show that at a duty cycle of 90\% or more, the area was sufficiently covered. As the duty cycle increased, the $R_p$ increased; and at too low a duty cycle, the desired area was not sufficiently covered. This not only results in insufficient flow but also does not reach the area with the corresponding flow rate. Therefore, it can be seen that the results according to the distance must be derived.

\subsubsection{Preliminary Experiment 2: Evaluation of Spraying Distance}
As the duty cycle was adjusted, the spray distance decreased. As the effective injection range decreases according to the duty cycle, the duty cycle value was set as a variable to optimize the duty cycle value by comparing $R_p$. An experiment was designed to find an appropriate duty cycle based on the depth data recognized by the camera. Preliminary experiment 2 was designed, as shown in Fig.~\ref{PE2_setup}, and the experiment was carried out by attaching a model tree to the mobile platform. The experiment was conducted with a total of six cases ranging from 75\% to 100\% based on the duty cycle. The distance was set according to the environment of the orchard in which the experiment was conducted. The depth between the nozzle and the target increased by 30cm, and the experiment was conducted from 70 to 160cm. 

The results of preliminary experiment 2, are shown in Fig.~\ref{PE_re2}. According to the results of preliminary experiment 2, at a duty of 90\% or more, the area was almost reached. At a close distance, it shows sufficient coverage, even at a low duty. Therefore, it is important to design an optimal duty cycle according to the distance to appropriately determine the amount of pesticide used to improve spraying performance.

\subsubsection{Preliminary Experiment 2: Evaluation of Spraying Distance}
\begin{figure}[!t]
\begin{center}
\includegraphics[width=14cm]{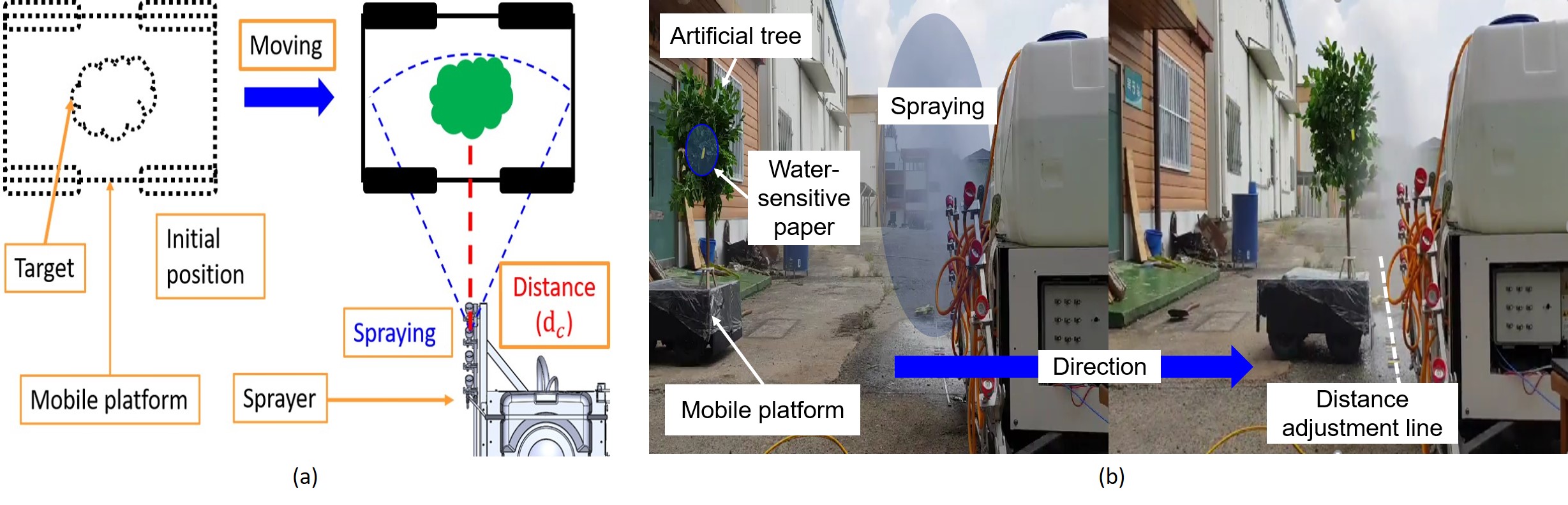}
\caption{Preliminary experiment 2 setup. (a) Configuration of preliminary experiment 2 for spraying performance evaluation. (b) Field view of the experiment.}
\label{PE2_setup}
\end{center}
\end{figure}

As the duty cycle was adjusted, the spray distance decreased. As the effective injection range decreases according to the duty cycle, the duty cycle value was set as a variable to optimize the duty cycle value by comparing $R_p$. An experiment was designed to find an appropriate duty cycle based on the depth data recognized by the camera. Preliminary experiment 2 was designed, as shown in Fig.~\ref{PE2_setup}, and the experiment was carried out by attaching a model tree to the mobile platform. The experiment was conducted with a total of six cases ranging from 75\% to 100\% based on the duty cycle. The distance was set according to the environment of the orchard in which the experiment was conducted. The depth between the nozzle and the target increased by 30cm, and the experiment was conducted from 70 to 160cm. 

\begin{figure*}[!t]

\begin{center}
\includegraphics[width=14cm]{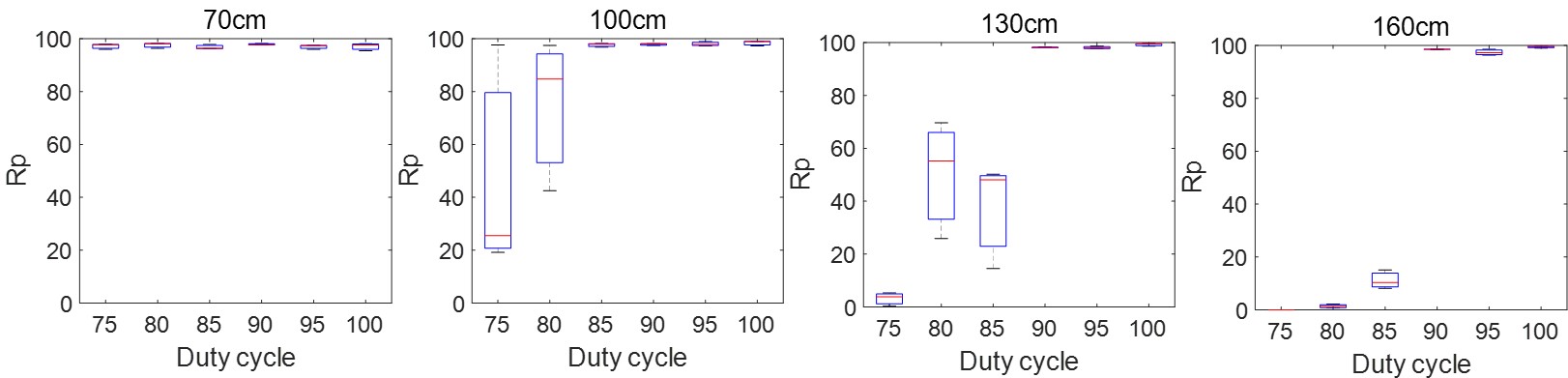}
\caption{Preliminary experiment 2 results.}\label{PE_re2}
\label{Preliminary experiment 2}
\end{center}
\end{figure*}

The results of preliminary experiment 2, are shown in Fig.~\ref{PE_re2}. According to the results of preliminary experiment 2, at a duty of 90\% or more, the area was almost reached. At a close distance, it shows sufficient coverage, even at a low duty. Therefore, it is important to design an optimal duty cycle according to the distance to appropriately determine the amount of pesticide used to improve spraying performance.

\subsection{Variable Flow Control Design}
The flow rate was determined by controlling the proportional solenoid valve. By controlling the plunger position of the proportional solenoid valve, the discharge flow rate was determined, and the flow rate could be controlled in real time. A proportional solenoid valve is achieved through the PWM control method, which is used to obtain a linear relationship between the fluid flow and the input signal using a proportional solenoid valve in a fluid power system~\cite{le2010comparison}.

\begin{figure}[!t]
\begin{center}
\includegraphics[width=12cm]{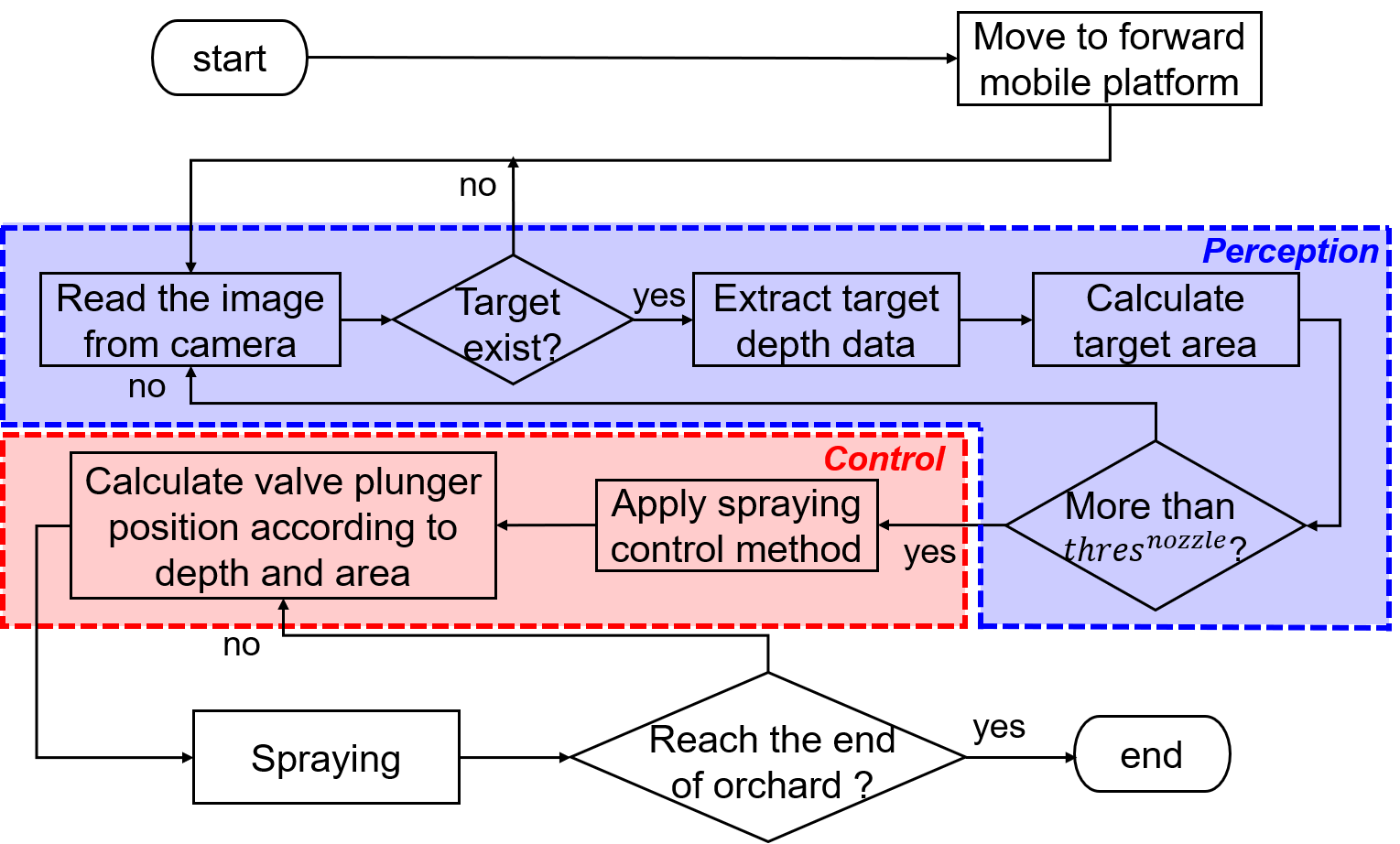}
\caption{Flow chart of intelligent spraying system.}\label{flowchart}
\end{center}
\end{figure}

\begin{figure}[!t]
\begin{center}
\includegraphics[width=12cm]{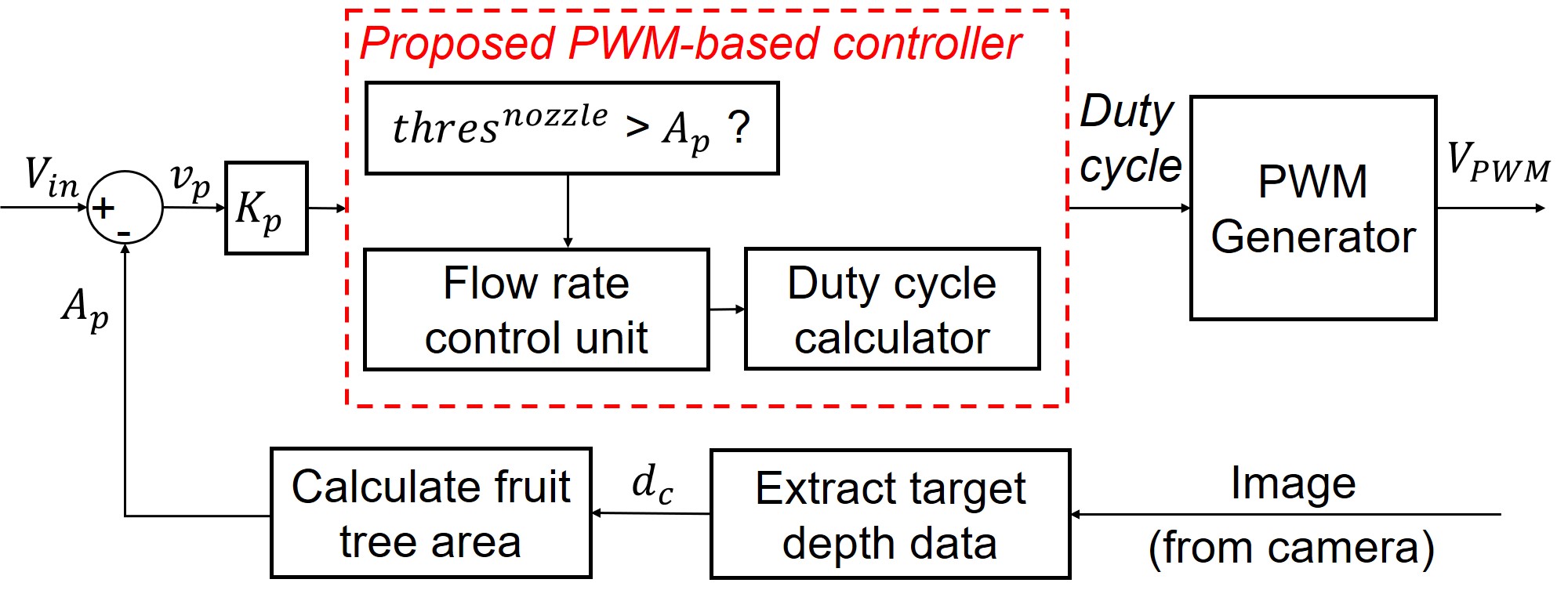}
\caption{Variable flow rate control diagram.}\label{control diagram}
\end{center}
\end{figure}

A flowchart of the intelligent spraying system is shown in Fig.~\ref{flowchart}. The flowchart provides an overview of the spraying procedure. A nozzle operation diagram is shown in Fig.~\ref{control diagram}. The image acquired from the camera, depth information of the fruit tree, and fruit tree area were calculated. At this time, the spray is determined according to whether it satisfies the threshold value set. $V_{PWM}$ is determined by the PWM-based controller, and the determined $V_{PWM}$ is generated through the PWM generator. The output signal $V_{PWM}$ is transmitted to the solenoid valve to adjust the plunger position.

The total flow rate Q is equal to the sum of the flow rates of each nozzle and is expressed as follows:

\begin{equation}
{Q}= Q_1 + Q_2 + ... + Q_8
\end{equation} 

The average flow rate of each proportional valve when the valve opens can be expressed as

\begin{equation}
{Q_{n}}= C_n \cdot A_n \cdot {x_{n}(t)} \cdot \sqrt{\frac{2\cdot P_n}{\rho}}
\end{equation} 

where {$C_n$} is the discharge coefficient, {$A_n$} is the throat area of the annular orifice, {$x_n(t)$} is the proportional valve plunger position, {$P_n$} is the pressure.

Each nozzle flow rate is controlled by the degree of opening and closing of the proportional valve. The degree of opening and closing of the proportional valve can be controlled according to the valve plunger position. The valve plunger position is related to the control voltage input $V_{PWM}$ and can be defined as

\begin{equation}
{{x_{n}(t)} = f(A_p,d_c,v_p,V_{PWM})}
\end{equation}

where {$A_p$} is the area of the fruit tree, {$d_c$} is the distance between the sprayer and the camera, {$v_p$} is the velocity of the platform, and {$V_{PWM}$} is the control input PWM duty cycle.

$V_{PWM}$ was determined using the control method. The control method is divided into on/off control and variable flow rate control in real time. In the case of on/off control, the following conditions must be satisfied to determine the on/off state of the valve. The state of the valve is determined by thres$^{nozzle}$ based on a predetermined threshold~\cite {kim2020intelligent}. In this study, we set this value to 10\%, and the control input $V_{PWM}$ is determined as follows according to the acquired fruit tree area: If the value is less than 10\%, then the valve is kept in the OFF state; and above it, the valve is maintained in the ON state.

\begin{equation}
{V_{pwm}} =\begin{cases} 0\% &  if \quad\quad  {A}_{p} \leq thres^{nozzle} \\100\% & else \end{cases} 
\end{equation}

where thres$^{nozzle}$ is the percentage of pixels that determines whether the nozzle is open.

The variable flow rate control in real time was controlled by a PWM signal to the flow rate control unit. The PWM control input is determined by $A_p$, $D_c$, $K_p$ and $C_v$. 

\begin{equation}
{V_{pwm}} =\begin{cases}\quad\quad\quad75$\%$ &  if \quad\quad d_c  \leq  0.9(m)\\{K_p} \times {A_p} \times {d_c} + {C_v}& else \end{cases} 
\end{equation}

where {$K_p$} is the proportional constant, and {$C_v$} is the dead zone according to valve dynamic characteristics. 

The variable flow rate control has the most influence on the control performance according to the design of the control input $V_{PWM}$. Based on the preliminary experiments, we optimized the control parameters. According to preliminary experiment 1, when the duty cycle was more than 90\%, the performance was the best, and there was no significant difference at the higher duty cycle. According to preliminary experiment 2, since the entire area is covered even at a low duty cycle at a distance of 0.9m or less, the minimum spraying amount is determined.

Based on the preliminary experiment, the flow rate was determined as shown in Eq.~6, and it is important to tune the proportional constant $K_p$ gain. Because the control is not carried out at less than 75\% due to the dynamic characteristics of the valve, it was adjusted so that the spray could be sprayed according to the distance and when it should be sprayed. At a close distance, all areas were well covered regardless of the spraying area, and the proportional gain value was adjusted according to the spraying distance. This was set to 0.8 through preliminary experiment 1 and 2.

\section{FIELD EXPERIMENTS}
\subsection{Experiment}

\begin{figure}[!t]
\begin{center}
\includegraphics[width=14cm]{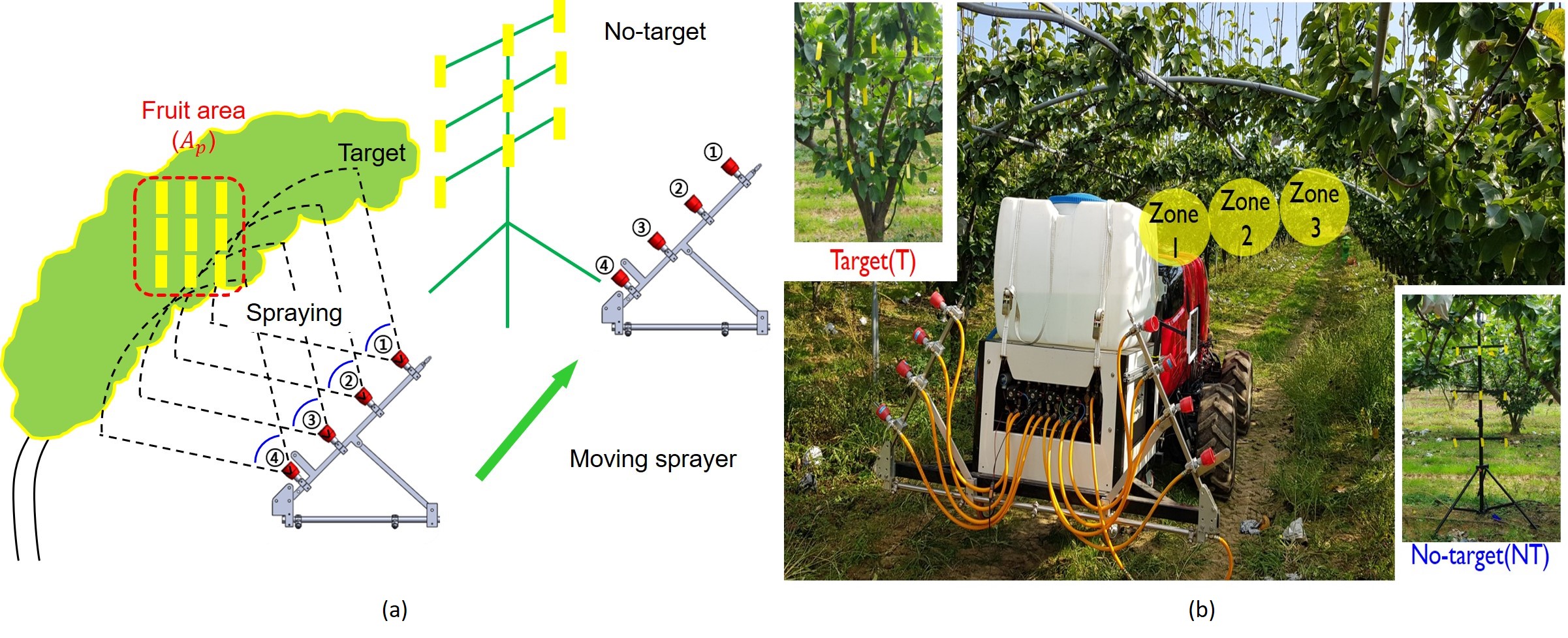}
\caption{Field experiment setup. (a) Configuration of field experiment for each control spraying performance evaluation. (b) Environment consisted of zone 3, where each zone was divided into target and no-target. Target is a should be sprayed because there is a fruit tree. Target tree is attached to nine water-sensitive papers. No-target does not require a spraying area because there is no fruit tree. No-target is attached to nine water-sensitive papers.}
\label{experiment setup}
\end{center}
\end{figure}

Field experiments on the variable spraying system were conducted at a pear orchard in Bonghwang, Naju, South Korea. As introduced in Section III-A, the variable spraying system was attached as an intelligent spraying system to a mobile platform, and the environment was configured as shown in Fig.~\ref{experiment setup}. A total of nine water-sensitive papers were attached by dividing into target (T) and no-target (NT) in a total of three zones. The 
T refers to an area where there is a fruit tree that must be sprayed, and NT refers to an area that does not need to be sprayed because there is no fruit tree.

The experiment was carried out with three controls:
\begin{enumerate}
    \item Control 1 -- All Nozzles Open (spraying without applying an intelligent spraying system) 
    \item Control 2 -- On/off Control (spraying while applying an intelligent spraying system)  
    \item Control 3 -- Variable Flow Rate Control (spraying while applying a variable spraying system)
\end{enumerate}

\subsection{Results}

Fig. 9--14 and Table II--V show each control spraying performance in the field experiment. Fig. 15 and Table. VI summarize each control spraying performance. The performance of three different experiments was analyzed by $R_p$ according to each control. Fig.~16--18 show snapshots of each control experiment. The differences occurring in each control are shown, and details are indicated in the caption of Fig.~16--18.

\subsubsection{Control 1 -- All Nozzles Open} The first experiment was conducted to confirm the degree of attachment to the water-sensitive paper when all the nozzles were opened.
 The intelligent control system can evaluate the spraying performance by comparing $R_p$ sprayed on the desired area, as in an experiment where control was not performed. In addition, it is possible to see how much the pesticide has decreased by comparing $R_p$ in the NT area, which is an undesired area, and by comparing the pesticide consumed to control the entire area, the degree to which the pesticide has decreased.
 
\begin{figure}[!t]
\begin{center}
\includegraphics[width=6.5cm]{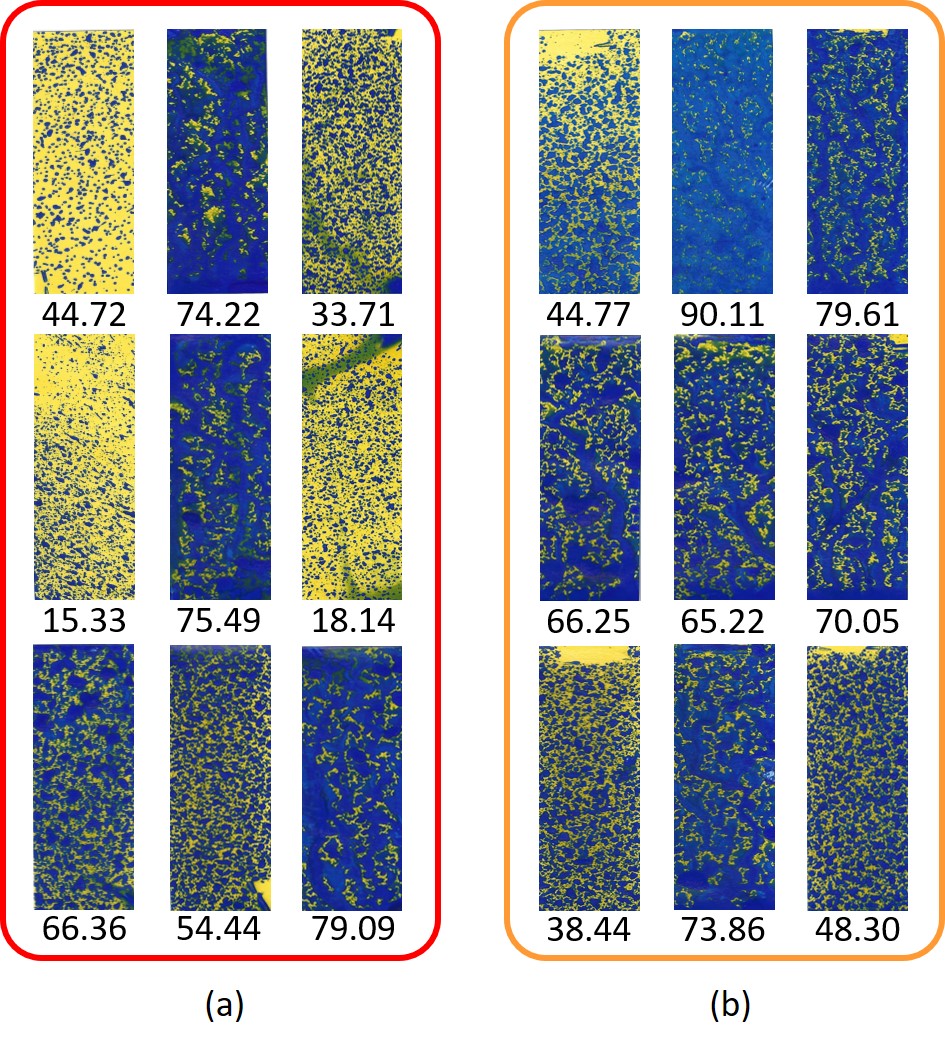}
\caption{Example of water-sensitive paper result in Control 1: All Nozzles Open; (a) Target, (b) No-target.}
\label{WSP_all}
\end{center}
\end{figure}

 T and NT in each zone $R_{p}$ are shown in Fig.~\ref{WSP_all}. The experiment was performed in two trials, and 54 water-sensitive papers used in the T and NT were used. It is assumed that this will be the same as in attempting 54 experiments with one water-sensitive paper attached. The amount of pesticide used in one row of orchards was 25 L. In the experiment, the $R_{p}$ value of the T and NT, which should be controlled, were recorded as an average of 56.15
 $\%$ and 58.80$\%$, respectively. These results show that $R_{p}$ is the same in all areas because the spraying is performed indiscriminately.

\begin{figure}[!t]
\begin{center}
\includegraphics[width=8cm]{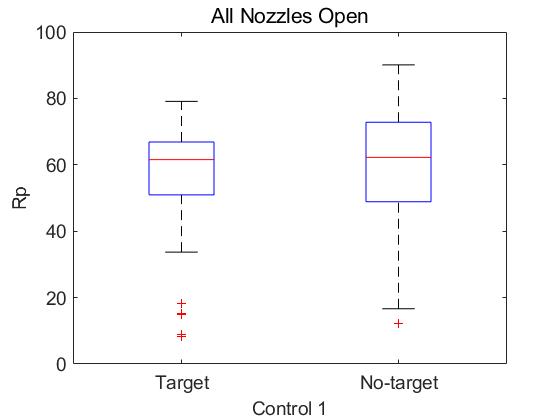}
\caption{Spraying result in Control 1 -- All Nozzles Open. Box plots display median, interquartile range, maximum, and minimum values. }\label{allopen}
\end{center}
\end{figure}

\begin{table}[!t]
\caption{Results of control 1: All nozzles open.}
\centering
\begin{tabular}{cc|ccc}
\hline
\multicolumn{2}{c}\textbf{\textbf{${R}_{p}$ (\%)}} & \textbf{Mean ($\pm$SD)} & \textbf{Max} & \textbf{Min} \\ 
\hline
\multirow{2}{*}{\textbf{Trial 1}} & \textbf{T} & \textit{53.27 (19.53)} & 72.37 & 14.85 \\
 & \textbf{NT} & \textit{58.88 (19.56)} & 89.77 & 12.25 \\ \hline
\multirow{2}{*}{\textbf{Trial 2}} & \textbf{T} & \textit{59.03 (14.40)} & 79.09 & 18.14 \\ 
 & \textbf{NT} & \textit{58.72 (13.95)} & 90.11 & 30.53 \\ 
\hline
\end{tabular}
\label{tab_1}
\end{table}

\subsubsection{Control 2 -- On/off Control} The second experiment was performed to confirm the degree of adhesion to the water-sensitive paper when spraying was performed by on/off control. In contrast to the results of the first experiment, the performance of the intelligent spraying system was obtained through an on/off control. If the result of the first experiment is the same for T, where the fruit tree exists, then it can be seen that the control in the desired area has been performed smoothly. In addition, if the result is lower than that of the first experiment in the NT area, which is an area where fruit trees do not exist, it can be seen that fewer pesticides existed in the area. These results show that the consumption of unnecessary pesticides is reduced.

\begin{figure}[!t]
\begin{center}
\includegraphics[width=6.5cm]{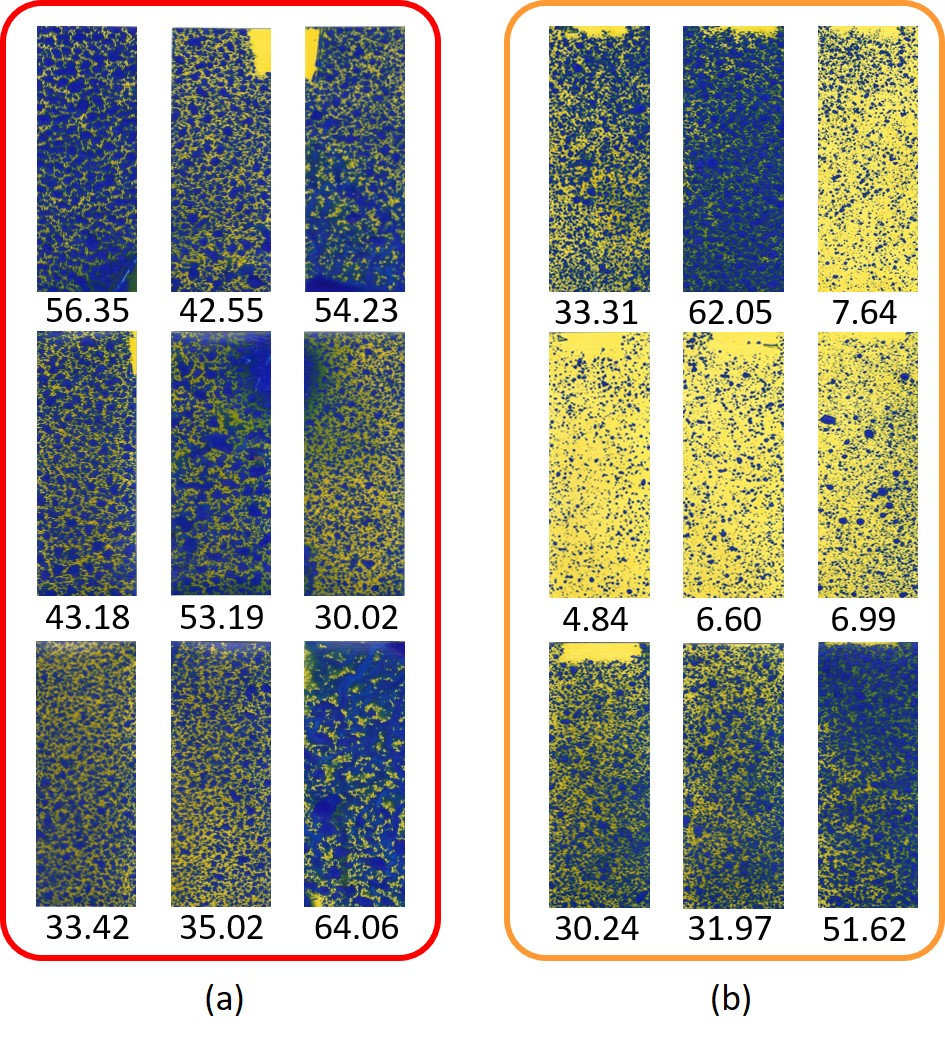}
\caption{Water-sensitive paper result in Control 2 -- On/off Control: (a) Target, (b) No-target.}

\label{WSP_on}
\end{center}
\end{figure}

The experimental results are presented in Fig.~\ref{WSP_on} and represent $R_{p}$ at T and NT in each zone, respectively. The amount of pesticide used in one row of orchard was 19.6 L. In the experiment, $R_{p}$ of the T and NT, which should be controlled, averaged 68.95$\%$ and 39.36$\%$, respectively. Compared with the first experiment, $R_{p}$ increased by 12.7$\%$ in T, and $R_{p}$ decreased by 18.44$\%$ in NT. As a result, it was shown that the $R_{p}$ did not decrease in the T, which is the area where the control needs to be performed, Thus, the control performance does not deteriorate, and the $R_{p}$ decreases in the NT, which is the area where control does not need to be carried out, reducing unnecessary pesticide use.

\begin{figure}[!t]
\begin{center}
\includegraphics[width=8cm]{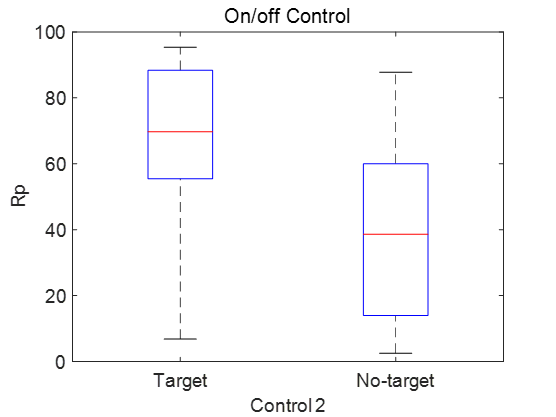}
\caption{Spraying result in Control 2 -- On/off Control. Box plots display median, interquartile range, maximum, and minimum values. }\label{onoff}
\end{center}
\end{figure}

\begin{table}[!t]
\caption{Results of Control 2: On/off Control.}

\centering
\begin{tabular}{cc|ccc}
\hline
\multicolumn{2}{c}\textbf{\textbf{${R}_{p}$ (\%)}} & \textbf{Mean ($\pm$SD)} & \textbf{Max} & \textbf{Min} \\ 
\hline
\multirow{2}{*}{\textbf{Trial 1}} & \textbf{T} & \textit{54.29 (17.40)} & 75.69 & 6.80 \\
 & \textbf{NT} & \textit{31.72 (22.14)} & 61.55 & 2.81 \\ \hline
\multirow{2}{*}{\textbf{Trial 2}} & \textbf{T} & \textit{83.60 (12.67)} & 95.36 & 48.77 \\ 
 & \textbf{NT} & \textit{47.01 (28.71)} & 87.78 & 6.69 \\ 

\hline
\end{tabular}
\label{tab_2}
\end{table}

\subsubsection{Control 3 -- Variable Flow Rate Control} A third experiment was carried out to confirm the degree of adhesion to water-sensitive paper when spraying was performed through variable flow rate control. This shows intelligent spraying system performance in contrast to the first experiment. In addition, it can be seen that the highest performance is shown by comparing the consumption of pesticides in the second experiment and showing the lowest consumption of pesticide.

The experimental results are shown in Fig.~13, 14 and represent the $R_{p}$ at T and NT in each zone. The amount of pesticide used in one row of orchard was 12.7L. In the test, the $R_{p}$ value of T and NT, which should be controlled, averaged 57.33$\%$ and 8.08$\%$, respectively. Compared to the first experiment, $R_{p}$ increased by 1.18$\%$ in the T, and $R_{p}$ decreased by 49.72$\%$ in NT. As a result, it was shown that $R_{p}$ did not decrease in the T, which is the area where the control needs to be performed, so that the control performance did not deteriorate; and $R_{p}$ decreased in the NT, the area where the control was not required. Thus, the amount of unnecessary pesticides was reduced. In addition, compared with the second experiment, this experiment showed a lower $R_{p}$ in the NT, and the amount of pesticide used was also low, which proved that this was the most effective control.

\subsubsection{Summary and discussion}
Fig.~15 and Table. V summarize the control results. Control 1 is representative of the sprayer performance because it performs conventional spraying. Therefore, the Control 1 result can serve as a reference. In the T, $R_p$ shows result of 56.15 (17.24)\%, 68.95 (21.12)\%, and 57.33 (21.73)\% for each control. Control 2 and 3 showed higher $R_p$ than Control 1. This indicates that the spraying performance does not decrease in the T. Control 2 is the highest $R_p$, and we believe that the high value was indicated by the difference in the pesticide form attached to the water-sensitive paper. Pressure differences in the on/off process appear to be the result of the derivation.

The NT shows results of 58.80 (16.83)\%, 39.37 (26.54)\%, and 8.08 (5.97)\% for each control. Control 2 and 3 showed lower $R_p$ than Control 1. This indicated that the undesired area was not sprayed. In particular, Control 3 significantly differed from Control 1, and was lower than Control 2. In this result, Control 3 spraying performance is more than Control 2 and indicates the highest performance. Therefore, the proposed Control 3 
spraying method was shown to be the most optimal control , and it can be seen that the flow rate modeling based on the preliminary experiment achieved optimization.

\begin{figure}[!t]
\begin{center}
\includegraphics[width=6.5cm]{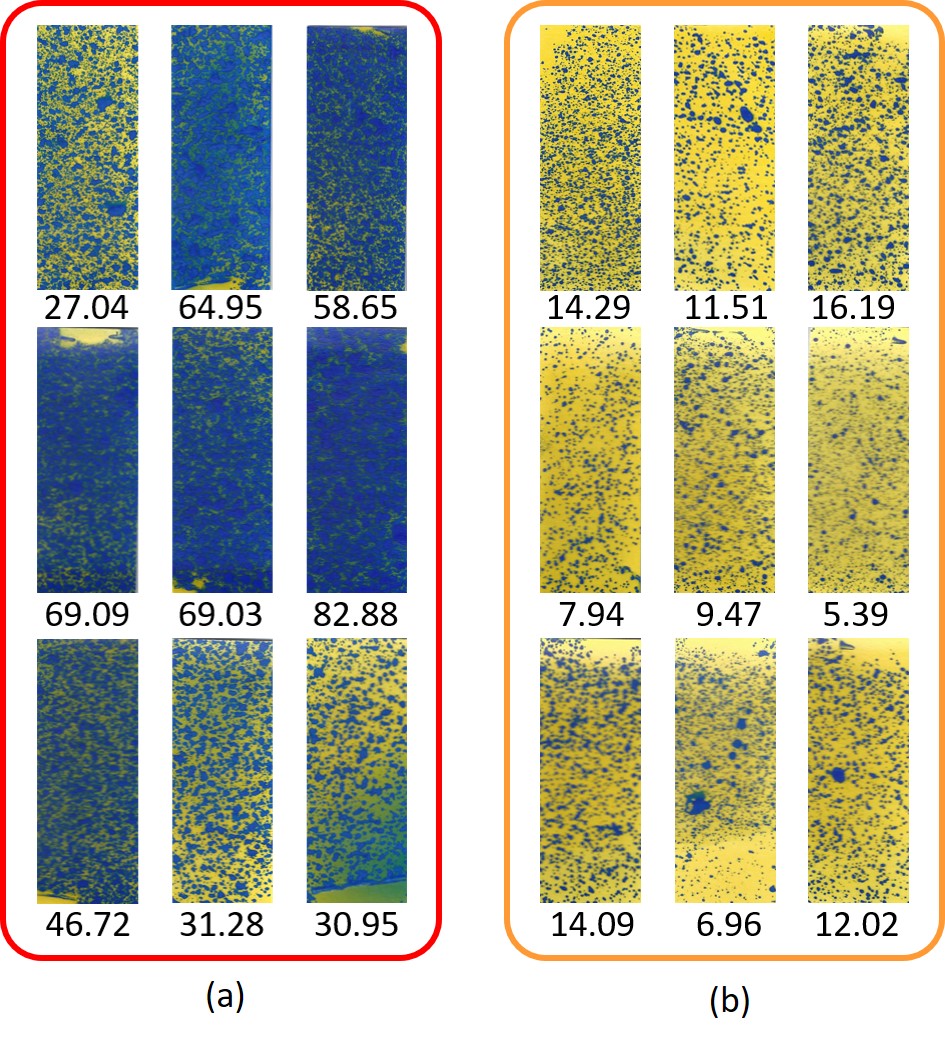}
\caption{Water-sensitive paper result in Control 3 -- Variable Flow Rate Control: (a) Target, (b) No-target.}

\label{Control3}
\end{center}
\end{figure}

\begin{figure}[!t]
\begin{center}
\includegraphics[width=8cm]{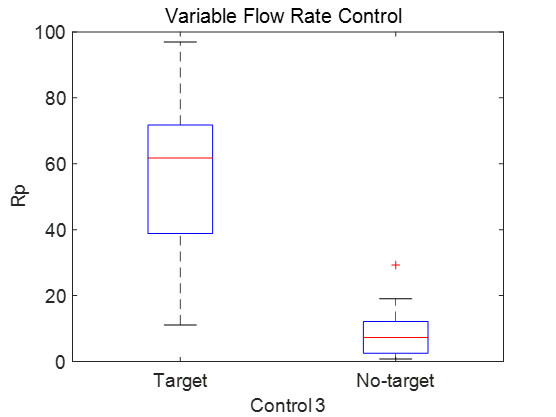}
\caption{Spraying result in Control 3 -- Variable flow rate control. Box plots display median, interquartile range, maximum, and minimum values. }\label{flowcontrol}
\label{Control3_1}
\end{center}
\end{figure}

\begin{table}[!t]
\caption{Results of Control 3: Variable flow rate control.}

\centering
\begin{tabular}{cc|ccc}
\hline
\multicolumn{2}{c}\textbf{\textbf{${R}_{p}$ (\%)}} & \textbf{Mean ($\pm$SD)} & \textbf{Max} & \textbf{Min} \\ 
\hline
\multirow{2}{*}{\textbf{Trial 1}} & \textbf{T} & \textit{54.43 (21.00)} & 84.45 & 19.92 \\
 & \textbf{NT} & \textit{10.63 (6.50)} & 29.25 & 0.77 \\ \hline
\multirow{2}{*}{\textbf{Trial 2}} & \textbf{T} & \textit{60.23 (22.46)} & 96.92 & 11.08 \\ 
 & \textbf{NT} & \textit{5.54 (4.11)} & 13.65 & 1.01 \\ 

\hline
\end{tabular}
\label{tab_3}
\end{table}

\begin{figure}[!t]
\begin{center}
\includegraphics[width=8cm]{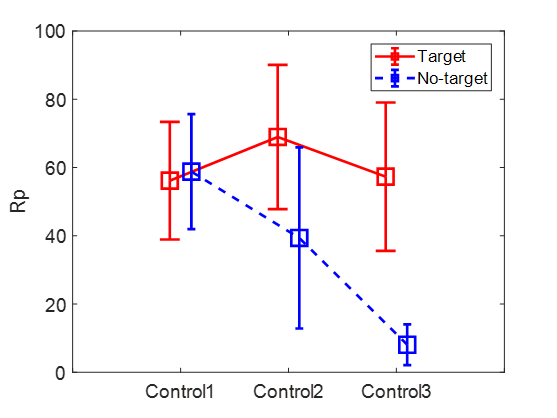}
\caption{Experimental results for each control. Figure is plotted as $mean \pm standard\: error.$  }\label{fig:1}
\end{center}
\end{figure}

\begin{table}[!t]
\caption{Results of field experiment}

\centering
\begin{tabular}{cc|ccc}
\hline
\multicolumn{2}{c}\textbf{\textbf{${R}_{p}$ (\%)}} & \textbf{Mean ($\pm$SD)} & \textbf{Max} & \textbf{Min} \\ 
\hline
\multirow{2}{*}{\textbf{Control 1}} & \textbf{T} & \textit{56.15  (17.24)} & 79.09 & 8.31 \\
 & \textbf{NT} & \textit{58.80 (16.83)} & 90.11 & 12.25 \\ \hline
\multirow{2}{*}{\textbf{Control 2}} & \textbf{T} & \textit{68.95(21.12)} & 93.74 & 6.80 \\ 
 & \textbf{NT} & \textit{39.37 (26.54)} & 83.19 & 2.51 \\ \hline
\multirow{2}{*}{\textbf{Control 3}} & \textbf{T} & \textit{57.33(21.73)} & 96.91 & 11.08 \\
 & \textbf{NT} & \textit{8.08 (5.97)} & 19.00 & 0.78 \\ 

\hline
\end{tabular}
\label{tab_4}
\end{table}

\section{DISCUSSIONS}
\subsection{Performance measure}
We proposed a deep learning-based intelligent spraying system and proved the superiority of variable flow control in real-time by comparing on/off control. To evaluate the spraying performance, it was confirmed that the performance was improved by comparing the pesticide adhesion rates of water-sensitive paper. The performance of the water-sensitive paper, may be seen as an insufficient result in that it represents only a specific area. Therefore, it is necessary to evaluate the system using a performance measure that is different from the pesticide adhesion rate.

As another example, in order to evaluate the performance of the water-sensitive paper, it will be possible to more accurately verify the spraying performance by analyzing the droplet size and pattern recorded on the water-sensitive paper. In this case, the water droplets formed on the leaf surface droop owing to gravity. The separation and analysis of these cases is a major issue.

\begin{figure*}[!t]
\begin{center}
\includegraphics[width=15cm]{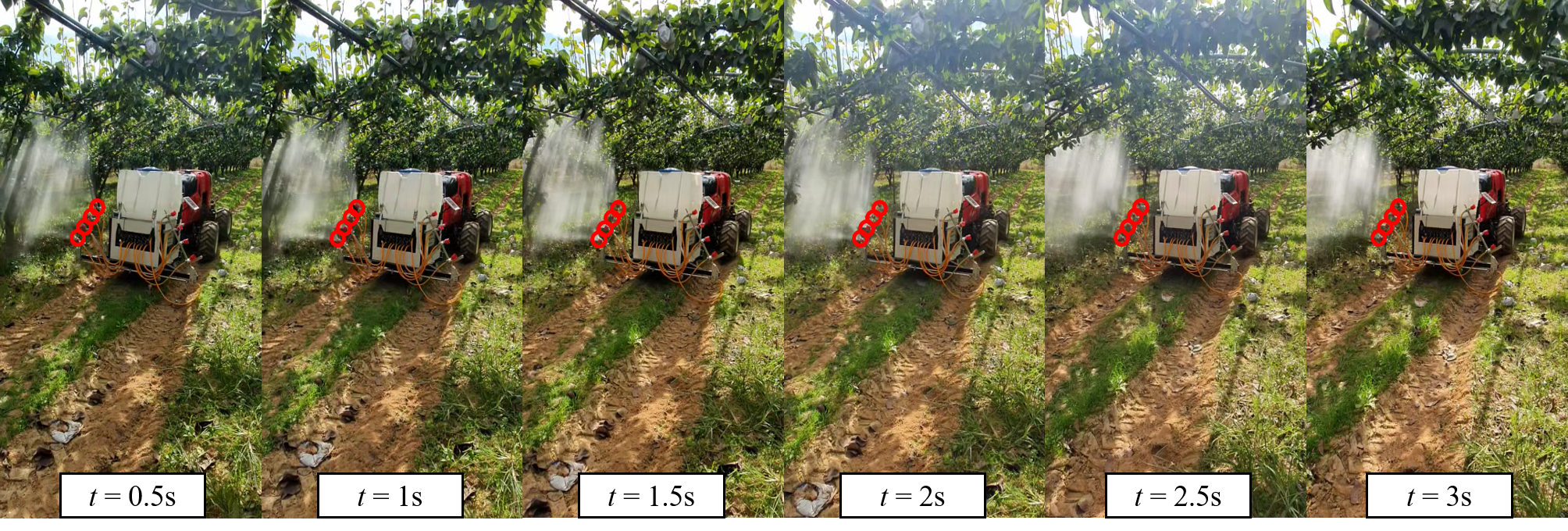}
\caption{Snapshots from 0.5 to 3 s of spraying. Control 1 was always open regardless of whether fruit tree was recognized. Each nozzle was always fully open, according to Control 1. Red circle indicates fully open nozzle. }\label{Ex_cap}
\end{center}
\end{figure*}

\begin{figure*}[!t]
\begin{center}
\includegraphics[width=15cm]{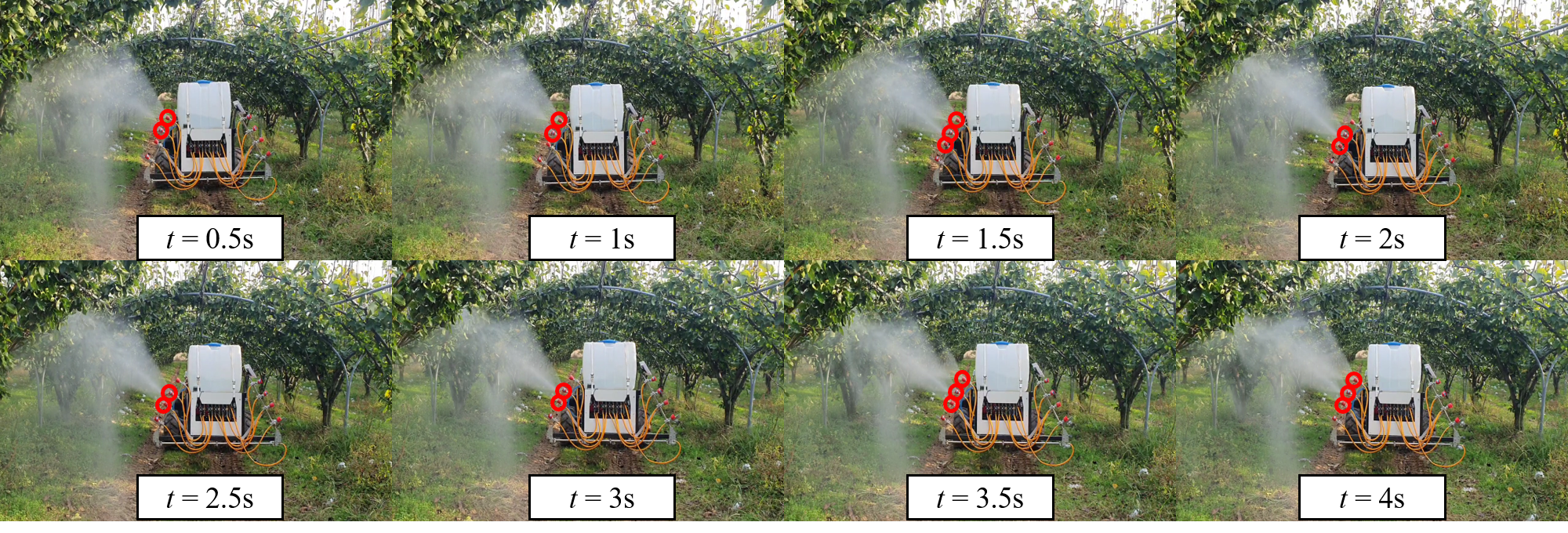}
\caption{Snapshots from 0.5 to 4s of spraying. Each nozzle was selective fully open nozzle for Control 2. When fruit tree area corresponding to each nozzle was recognized, then nozzle was fully open. Red circle represent a fully open nozzle. For example, when t = 1 s, two nozzles were opened because area corresponding to top two nozzles was recognized as a fruit tree; and when t = 1.5 s, area corresponding to top three nozzles was recognized as a fruit tree and three nozzles were opened. }\label{Ex2_cap}
\end{center}
\end{figure*}

\begin{figure*}[!t]
\begin{center}
\includegraphics[width=15cm]{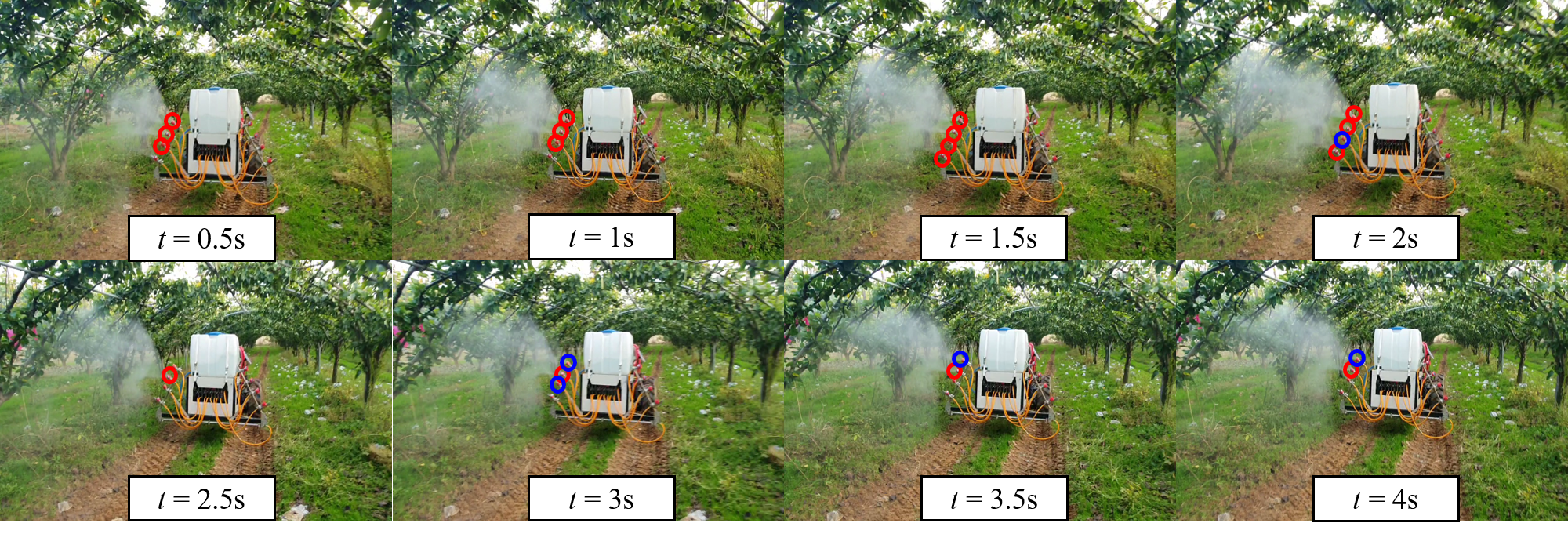}
\caption{Snapshots from 0.5 to 4 s of spraying. Each nozzle flow rate was controlled by Control 3. Flow rate of each nozzle was determined by calculating area and distance of fruit tree in area corresponding to each nozzle. For example, when t = 2 s, all four nozzles were open, but compared to t = 1.5 s, third nozzle from top smallest spraying. When t = 3.5 and 4 s, amount of spraying at first nozzle was significantly reduced. Red circle represents a fully open nozzle. Blue circle indicates  nozzle not fully open according to variable flow rate control. }\label{Ex3_cap}
\end{center}
\end{figure*}

\subsection{Precision level}
Currently, it is thought that a very precise level of spraying is not necessary in fruit trees, but as precision agriculture develops in the future, the importance of precise agriculture in the open field and orchards may be highlighted. Therefore, it is necessary to conduct a study on the analysis of how much the pesticide has reached by sensing the spraying type and droplet, and such a study will be related to drift.

However, it is expected that it will be very difficult to control the precise level of spraying owing to the effects of drift. For precise spraying, a method to suppress drift should be researched. If a study to minimize drift by changing the structure of the nozzle and a study to analyze the degree of drift according to the pump pressure are carried out, then the control parameters can be optimized.

\subsection{Drift analysis}
As one of the factors that has the greatest influence on the spraying performance, drift can cause damage to other crops. However, the drift is mainly caused by wind, therefore it is difficult to analyze accurately. In the case of the proposed intelligent spraying platform, there is no wind assist device used in other orchards, so it is expected that drift will not occur over long distances. However, in the case of drift, it is difficult to control the drift because external forces such as wind cannot be controlled. Therefore, measuring the drift and see determining the effect of drift is a challenge.

It is possible to investigate the distance and effect of drift by investigating the overlapping pesticides as an indirect measurement method, but it is not known exactly what variables are required. However, if the drift can be precisely sensed through a vision sensor, then the factors that affect the spray can be immediately confirmed visually confirmed. If the factors are accurately confirmed, then policies to minimize drift can be established. However, crops that exist in an open field environment or an orchard environment can act as obstacles to drift sensing.


\section{CONCLUSIONS}
In this paper, we proposed a variable flow control system in real time with deep-learning-based fruit tree perception. Theoretical modeling may differ because the pressure acts as a variable on the actual nozzle tip. Flow rate modeling was designed by a preliminary experiment that evaluated whether the result completely covers the spraying area according to the duty cycle. An evaluation of spraying distance examined whether the output reached the corresponding spraying area according to the duty cycle. Field experiments were conducted for three controls, and the results of each experiment confirmed that the pesticide was reduced in areas where spraying should not be performed at 56.80\%, 39.37\%, and 8.08\%. In addition, the actual use of pesticides decreased to 25, 19.6, and 12.7 L in the experiment, confirming that the proposed variable flow rate control system is more effective than the existing control method. We discussed the performance measures, research direction, and drift analysis.

\section*{ACKNOWLEDGMENT}

This research was supported, in part, by the Korea Institute for Advancement of Technology (KIAT) grant funded by the Korea Government (MOTIE) (P0008473, HRD Program for Industrial Innovation); in part, by Korea Institute of Planning and Evaluation for Technology in Food, Agriculture and Forestry(IPET) through Agriculture, Food and Rural Affairs Convergence Technologies  Program for Educating Creative Global Leader Program, funded by Ministry of Agriculture, Food and Rural Affairs(MAFRA)(716001-7).


\bibliographystyle{unsrtnat}
\bibliography{references}  

\end{document}